\documentclass[letterpaper, 10 pt, conference]{ieeeconf}  % Comment this line out if you need a4paper
\pdfoutput = 1

\IEEEoverridecommandlockouts                              % This command is only needed if 
\usepackage{amsmath,amssymb,amsfonts}
\usepackage[dvipdfmx]{graphics}
\usepackage{epsfig} % for postscript graphics files
\usepackage{cite}
\usepackage{algorithm}
\usepackage{algorithmic}
\usepackage{array}
\usepackage[caption=false]{subfig}
\usepackage{stfloats}
\usepackage{url}
\usepackage{enumerate}
\usepackage{color}
\usepackage{xcolor}
\usepackage{subfiles}
\usepackage{comment}
\usepackage{bm}
\usepackage{bbm}
\usepackage{kantlipsum}

\newtheorem{theorem}{Theorem}

\newtheorem{remark}{Remark}

\newtheorem{proposition}{Proposition}
\newtheorem{definition}{Definition}

\newtheorem{example}{Example}

\newcommand{\mbr}{\mathbb{R}}

\title{\LARGE \bf
Energy-Aware Task Allocation for Teams of Multi-mode Robots}

\author{Takumi Ito$^{1}$, Riku Funada$^{1}$, Mitsuji Sampei$^{1}$, and Gennaro Notomista$^{2}$% <-this % stops a space
\thanks{*This work was supported by JSPS KAKENHI, Grant Number 21H01348 and JST SPRING, Grant Number JPMJSP2180 (T. Ito, R. Funada, M. Sampei), and by the NSERC Discovery Grant RGPIN-2023-03703 (G. Notomista).}% <-this % stops a space
\thanks{$^{1}$ Department of Systems and Control Engineering, Institute of Science Tokyo, Tokyo 152-8550, Japan, {\tt\small  takumi.ito@sl.sc.e.titech.ac.jp, \{funada, sampei\}@sc.e.titech.ac.jp.}}%
\thanks{$^{2}$ Department of Electrical and Computer Engineering, University of Waterloo, Waterloo, ON N2L 3G1, Canada, {\tt\small  gennaro.notomista@uwaterloo.ca.}}%
}

\begin{document}

\maketitle
\thispagestyle{empty}
\pagestyle{empty}

%%%%%%%%%%%%%%%%%%%%%%%%%%%%%%%%%%%%%%%%%%%%%%%%%%%%%%%%%%%%%%%%%%%%%%%%%%%%%%%%
\begin{abstract}
This work proposes a novel multi-robot task allocation framework for robots that can switch between multiple modes, e.g., flying, driving, or walking. We first provide a method to encode the multi-mode property of robots as a graph, where the mode of each robot is represented by a node. Next, we formulate a constrained optimization problem to decide both the task to be allocated to each robot as well as the mode in which the latter should execute the task. The robot modes are optimized based on the state of the robot and the environment, as well as the energy required to execute the allocated task. 
Moreover, the proposed framework is able to encompass kinematic and dynamic models of robots alike. Furthermore, we provide sufficient conditions for the convergence of task execution and allocation for both robot models.
\end{abstract}
%%%%%%%%%%%%%%%%%%%%%%%%%%%%%%%%%%%%%%%%%%%%%%%%%%%%%%%%%%%%%%%%%%%%%%%%%%%%%%%%
\section{INTRODUCTION}

Multi-robot systems perform increasingly complex tasks in numerous application fields, such as environmental monitoring \cite{10527388}, rescue \cite{saeedvand_robust_2019}, and delivery \cite{yakovlev2020delivery}. As such applications expand, today's growth in robot technology raises the variety and complexity of robotic systems. 
% Heterogeneous multi-robot systems have enhanced variation and resilience in tasks because if one robot is unable to execute a task, another robot can take its place. 
Many robots are now designed with switchable operation or locomotion modes, such as flying, driving, and walking \cite{DUCARD2021107035, lee_learning_2024, sihite_multi-modal_2023}. This multi-mode property makes robots more flexible, scalable, and resilient.

Switching between multiple modes provides advantages in both task execution and energy consumption. For instance, the convertible Unmanned Aerial Vehicle (UAV) surveyed in \cite{DUCARD2021107035} has two flight modes: cruise and hovering. Cruise mode is energy-efficient in high forward velocity while hovering mode has advantages in takeoff spaces and static hovering. Similarly, robots with wheels and legs \cite{lee_learning_2024}, or UAVs with wheels \cite{sihite_multi-modal_2023}, are more energy efficient when using wheels. However, walking or flying offers advantages in executing tasks on harsh terrain. 
%in a robot with both wheels and legs \cite{lee_learning_2024} or in a UAV with wheels \cite{sihite_multi-modal_2023}, locomotion with wheels is energy efficient. The other mode, i.e., walking or flying, has advantages in task execution under bad terrain conditions. 

In a scenario where multiple robots execute multiple tasks, Multi-Robot Task Allocation (MRTA) methods play a central role (see, e.g., \cite{gerkey2004mrtataxonomy, korsah2013taxonomy}). %(see, e.g., the taxonomies in \cite{gerkey2004mrtataxonomy, korsah2013taxonomy}). 
One key factor of MRTA is \textit{energy}. Many MRTA problems were formulated to minimize travel distance \cite{8862941} or energy consumption \cite{gennaro2022resilient}, or to manage limited energy resources \cite{7833258}. These approaches can be regarded as energy-saving strategies. 
Another important factor of MRTA is \textit{resilience}. Heterogeneous multi-robot systems enhance resilience since if one robot fails, another can take over. The robots' heterogeneity is encoded in \cite{ravichandar_strata_2020} and \cite{prorok2017impact}. Our previous work \cite{gennaro2022resilient} expanded to address component failures and facilitate task reallocation.
The robot's multimodality has the potential to enhance both factors by selecting the most suitable mode according to capability and energy consumption. Nevertheless, to the best of the authors' knowledge, no studies within the context of MRTA have explicitly considered the heterogeneity of multi-robot systems resulting from the multimodality of robotic units.

Numerous approaches have accomplished MRTA, including \textit{market-based approaches} \cite{quinton_market_2023} and \textit{behavior-based approaches} \cite{681242}. 
In contrast to many approaches, our previous work \cite{8795895} formulated an \textit{optimization problem} where each robot performs multiple tasks simultaneously with prioritizing, which is highlighted by lower computational complexity and dynamic allocation. Furthermore, we provided theoretical guarantees to the convergence of task execution and stable allocation in \cite{gennaro2022resilient}.
% However, many existing approaches, including \cite{quinton_market_2023,681242}, do not account for \textit{online} task allocation. 
%However, \textit{online} task allocation is not achievable with many existing approaches. 
% As noted in \cite{CHAKRAA2023104492}, optimization-based strategies have advantages in both performance and computational complexity. %, and their results are deterministic. 
Moreover, when the optimization problem is convex, the formulation is scalable with respect to the number of robots and tasks (see, e.g., the constrained-based task execution approach in \cite{9196741} based on Control Barrier Functions (CBFs) \cite{robot_ecology}, which is, however, limited to velocity-controlled robots).

% For these reasons, in this work we opt for an optimization-based allocation of tasks to multi-mode robots.

% Heterogeneity in multi-robot systems has also been studied in the MRTA field. Heterogeneous multi-robot systems have enhanced variation and resilience in tasks because if one robot is unable to execute a task, another robot can take its place. In \cite{parker1994heterogeneous}, Parker produced a method for allocating tasks to heterogeneous robots based on predefined behavioral habits. The relations between different robot species and abilities (traits) were explicitly encoded in \cite{ravichandar_strata_2020} and \cite{prorok2017impact}. Then, our previous work \cite{gennaro2022resilient} extended the encoding framework into a suitable form to be resilient to component failures on robots and changes in environmental situations. Nevertheless, to the best of the authors' knowledge, no studies within the context of MRTA have explicitly considered the heterogeneity of multi-robot systems resulting from the multimodality of robotic units.

This letter proposes a novel task allocation framework that can switch between multiple modes of robots. 
% To optimize the operations of teams of multi-mode robots, we focus on the following objectives: modes must be dynamically prioritized based on its capability and energy consumption, and task assignment to the modes needs to be accomplished online. 
The multimodality concept can be interpreted as a form of heterogeneity, and this analogy can be used to encode it. However, encompassing robots with multiple modes in task allocation frameworks is not straightforward. We further consider prioritizing between robot modes not only based on their capability, but also accounting for the energy consumption during task execution.%, and restricting overlap task assignment to multiple modes of the same robot.
% Hence, we employ the optimization-based approach, which has advantages in online computation and scalable constraints.

% Most existing works on task allocation and execution consider velocity-controlled robots and formulate task execution with the above-mentioned CBF-based approach \cite{robot_ecology}.
Although most existing works formulate task execution with velocity-controlled robots with the %above-mentioned 
CBF-based method \cite{robot_ecology}, many real robots have dynamics with high relative degree (e.g., acceleration-controlled robots executing position-related tasks). This makes formulating task executions difficult and requires a more sophisticated approach, such as integral CBF \cite{9131819} or cascaded CBF \cite{robot_ecology}. 
% Compared to state-of-the-art optimization-based task allocation frameworks, 
Our proposed approach is able to encompass tasks and robot models, which result in high relative degree constraints. We theoretically examine a convergence condition of the task execution and allocation in both kinematics and high relative degree cases.

To summarize, the contributions of our work can be written as follows:
1) We propose a new MRTA framework that can handle robots with multiple modes.
2) We propose task execution strategies that can be applied to robot and task models resulting in high relative degree constraints.
3) We provide theoretical conditions for convergence of task execution and allocation, considering both kinematic and dynamic robot models.
% \begin{enumerate}
%     \item We propose a new MRTA framework that can handle robots with multiple modes
%     \item Additionally, the proposed framework can be applied to robots and task resulting in high relative degree constraints
%     \item We provide theoretical conditions for convergence of task execution and allocation, considering both kinematic and dynamic robot models
% \end{enumerate}

\section{MULTI-MODE ROBOT MODELING}
 
\subsection{Task-to-mode Mapping}

\begin{figure}
    \centering
    \includegraphics[width=0.9\linewidth]{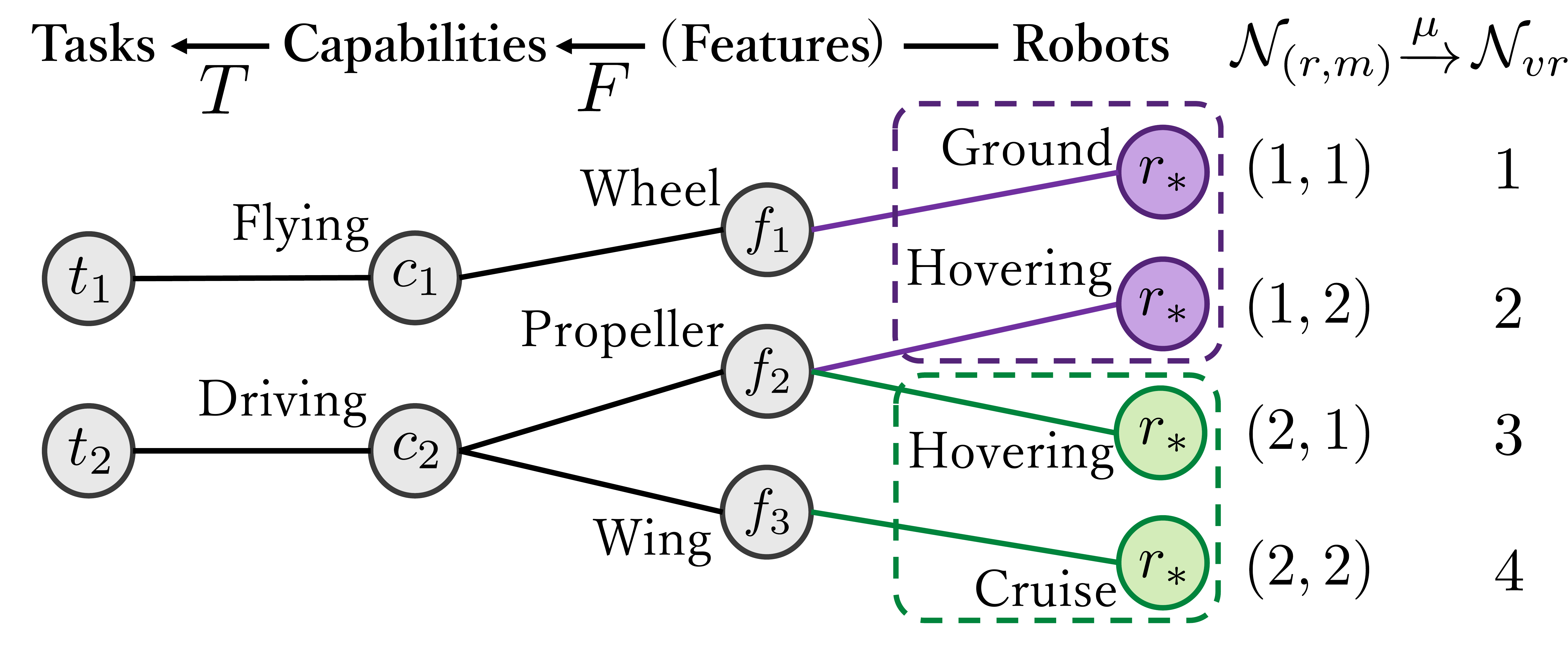}
    \caption{Example of mapping, including two tasks, two capabilities, three features, and two robots with two modes. One robot is a UAV with wheels, which has ground vehicle and ground modes, and the other is a convertible UAV, which has hovering and cruise modes.}
    \label{fig:multimode_encode}
\end{figure}

% This research aims to construct the framework that encodes the robot's multiple modes. 
The overview of the proposed encoding mapping is depicted in Fig.~\ref{fig:multimode_encode}. 
% Let the robot have multiple switchable modes. For instance, a convertible UAV has two flight modes: cruise and hovering, and a drone with wheels has two modes: aerial and ground operation. 
An encoding framework for robot heterogeneity is proposed in \cite{gennaro2022resilient} with multiple layers: robots, features, capabilities, and tasks. 
The layer of robots incorporates two factors: heterogeneity of robots (distinguished by purple and green colors in Fig.~\ref{fig:multimode_encode}) and modes of each robot (nodes in each of the dashed boxes). Each node expressing a mode is denoted by a \textit{virtual robot} hereafter. 

Let us exemplify how this framework incorporates a scenario in which a wheel-equipped UAV, which is depicted as purple nodes in Fig.~\ref{fig:multimode_encode}, conducts two tasks: 1) monitoring a targeted area from the air and 2) traversing a region not allowed flying due to law regulations. Each task requires a robot to own a specific capability. This requirement 
%associated with each task i
is expressed by a connection between the nodes of tasks and capabilities. Similarly, each capability is connected by features that grant a robot its capability. A wheel-equipped UAV switches its mode between hovering and ground by selecting a virtual robot according to its features.
% These features switch once a wheel-equipped UAV changes its mode between hovering and ground vehicle mode; each mode is expressed by a virtual robot in the layer of robots.

%Robots equip their features, several features attain capabilities, and some capabilities execute a task (e.g., see later Example \ref{ex:wheeled_UAV} and \ref{ex:convertible_UAV}). 
% The concepts of multiple modes and heterogeneity are similar. We can utilize this analogy to encode multi-mode; a switching mode can be regarded as switching its features. Hence, we set a \textit{virtual robot}, which has features of each mode of each robot.

Let the number of robots, features, capabilities, and tasks be $n_r$, $n_f$, $n_c$, and $n_t$, respectively.
% \red{(see \cite{gennaro2022resilient} for distinction of them) move later explain}. 
The following index sets are subsets of natural number $\mathbb{N}$.
The ordered sets of robot and task indices are denoted as $\mathcal{N}_r=\{1,\dots,n_r\}$ and $\mathcal{N}_t=\{1,\dots,n_t\}$.
%Additionally, 
Let robot $i$'s number of modes and a set of its indices be $m_i$ and $\mathcal{N}_{m_i}=\{1,\dots,m_i\}$, respectively. 

From the robot and mode indices, we now define a set of ordered pairs of robot and mode as $\mathcal{N}_{(r,m)}=\{(1,1),\dots,(1,m_1),\dots,(n_r,1),\dots,(n_r,m_{n_r})\}$, whose number of elements is denoted as $n_{vr}=\sum_{i\in\mathcal{N}_r}m_i$. 
Then, we define a set of virtual robot indices as $\mathcal{N}_{vr}=\{1,\dots,n_{vr}\}$, which is order isomorphic to $\mathcal{N}_{(r,m)}$.
Let us denote ``virtual robot of robot $i$ in mode $k$'' as ``virtual robot $\mu(i,k)$'' with a bijection mapping $\mu:\mathcal{N}_{(r,m)}\to\mathcal{N}_{vr}$.
For instance, the mapping $\mu$ in the case of two robots with two modes is shown in the right part of Fig.~\ref{fig:multimode_encode}. 
% A robot-mode pair index is mapped into a unique index.
% In this paper, we assume all robots have the same number of modes. However, this framework can be readily extended to allow for a flexible number of modes for each robot.
Now, we can leverage the robot-to-task mapping introduced in \cite{gennaro2022resilient}, using virtual robots, as explained in the following.

\subsubsection{Mapping Matrices}

% First, let us define a \textit{robot-to-feature mapping} as 
% \begin{align}
%     A \in \{ 0,1\} ^{n_f \times n_{vr}},\label{eq:robot_to_feature}
% \end{align}
% where $A_{\ell,\mu(i,k)}=1$ if and only if mode $k$ of robot $i$ possesses feature $\ell$. 
First, let us define a \textit{robot-to-capability mapping} matrix as $F \in \{ 0,1\} ^{n_c \times n_{vr}}$, where $F_{\ell,\mu(i,k)}=1$ if and only if virtual robot $\mu(i,k)$ possesses capability $\ell$. Note that this robot-to-capability mapping is formally defined using \textit{robot-to-feature mapping} and a \textit{feature-to-capability mapping} defined as bipartite hypergraph; refer to \cite{gennaro2022resilient} for a complete definition.
Then, a \textit{capability-to-task mapping} matrix is defined as $T\in \{ 0,1\} ^{n_t \times n_c}$, where $T_{j,\ell}=1$ if and only if task $j$ requires capability $\ell$.

\subsubsection{Specialization and Penalty}\label{sss:specialization}
Additionally, we define the \textit{specialization matrix}, which encodes potential candidates of tasks for virtual robots, as $S_{\mu(i,k)} = \mathrm{diag}(\mathbbm {1}_{n_t}-\mathrm{kron}(T F_{-,\mu(i,k)}))\in \mathbb {R}^{n_t\times n_t}$, where $\mathbbm {1}_{n_t}\in\mbr^{n_t}$ is a vector consisting of $1$, and $\mathrm{kron}$ is the Kronecker delta function. 
The symbols $X_{i,-}$ and $X_{-,j}$ denote the $i$th row and the $j$th column of the matrix $X$, respectively.
The specialization matrix $S_{\mu(i,k)}$ becomes a diagonal matrix whose $(j,j)$ element is $1$ if task $j$ is a potential candidate for virtual robot $\mu(i,k)$; otherwise, it is $0$. Specifically, when virtual robot $\mu(i,k)$ possesses at least one capability to support task $j$, the element value becomes $1$. 
Moreover, we define a penalty matrix as $\Pi_{\mu(i,k)} = I_{n_t} - S_{\mu(i,k)} S_{\mu(i,k)}^\dagger$, where the dagger is pseudo inverse.
The penalty matrix can be regarded as a penalty for task assignment to an infeasible robot since its $(j,j)$ element has a nonzero value if the task $j$ is infeasible, contrasting the specialization matrix. 

%\begin{example}[Robot 1 in Fig.~\ref{fig:multimode_encode}]
%\label{ex:wheeled_UAV}
%    Consider a wheel-equipped UAV with hovering and ground vehicle modes. Each mode has different features that support different capabilities. This robot can execute different tasks requiring different capabilities by switching its modes.
%\end{example}

\begin{example} \label{ex:convertible_UAV}
    Consider a convertible UAV with cruise and hovering modes, which are depicted as green nodes in Fig.~\ref{fig:multimode_encode}. The virtual robots of cruise and hovering modes have \textit{wing} and \textit{propeller}, respectively, and both features support \textit{flying} capability. Although both modes can execute the same task, energy efficiency can be differentiated by specifying distinct energy costs, which are introduced later. 
    By assigning a lower energy cost to cruise mode at high forward velocities than hovering mode, we can enable cruise mode to excel in tasks requiring high forward speed.
    This robot can adjust energy efficiency by switching its modes while executing the same task.
\end{example}

\subsection{System Models}
\subsubsection{Motion Model}
Consider the virtual robot $\mu(i,k)$.
Let the dimension of the state and input be $n_{x_i}$ and $n_{u_{i,k}}$, respectively.
Let the state of the robot be shared for all modes; then, we denote the state only with a subscript $i$ as $x_i\in\mbr^{n_{x_i}}$. In contrast, the input is denoted as $u_{i,k}\in\mbr^{n_{u_{i,k}}}$. In this letter, we consider mechanical robotic systems, therefore we model their dynamics via control affine dynamical systems. The state evolution of virtual robot $\mu(i,k)$ is then governed by the following differential equation:
\begin{align}
    \dot{x}_i = f_{i,k}(x_i) + g_{i,k}(x_i)u_{i,k},\label{eq:dynamics}
\end{align}
where $f_{i,k}: \mbr^{n_{x_i}}\to\mbr^{n_{x_i}}$ and $g_{i,k}: \mbr^{n_{x_i}}\to\mbr^{n_{x_i}\times n_{u_{i,k}}}$ are locally Lipschitz continuous.

\subsubsection{Energy Model}
The above discussion can distinguish the capability of task execution of each mode. However, if the capability is the same for multiple modes, how can we allocate tasks? Here, we introduce the energy cost to differentiate the task execution performance based on the energy consumption.

% The existing work reduces energy consumption by minimizing $\|u\|^2$. However, this does not necessarily mean minimizing energy. 
The previous works (e.g., \cite{gennaro2022resilient}) represent energy cost as $\|u\|^2$. This can be more generalized for some situations.
For instance, an airplane is operated more energy efficiently at high speeds than at low speeds because it can utilize aerodynamic lift. 
Therefore, in this letter, we consider the energy cost function a more general convex function as
\begin{align}
    &\varepsilon_{i,k}(u_{i,k})= \|u_{i,k} - u^\text{eff}_{i,k}\|^2_{w_{i,k}}, \label{eq:general_energy_cost}
\end{align}
where $w_{i,k}$ is a weight for the norm, and $u^\text{eff}$ is the input that minimizes energy cost (e.g., specific forward velocity). 
This formulation does not necessarily represent exact energy consumption. However, the quadratic formulation can approximate it and has advantages in handling and implementation. 

% \begin{remark}\label{rmk:energy_cost}
%     The existing work reduces energy consumption by minimizing $\|u\|^2$. However, this does not necessarily mean minimizing energy. For instance, an airplane is operated more energy efficiently at high speeds than at low speeds because it can utilize aerodynamic lift. Therefore, in this paper we consider as energy cost function a more general convex function of $u$, denoted by $\varepsilon(u)$.
% \end{remark}

\section{TASK EXECUTION AND ALLOCATION} \label{sec:task_allocation}
 
\subsection{Preliminary: Constraint-Based Task Execution} \label{sss:task_as_constraint}

This subsection introduces a multiple-task execution method based on the extended set-based tasks \cite{9196741}. That can theoretically guarantee the convergence of the task completion.
For the following discussion, we will temporarily omit the subscripts of the indices of robots, modes, and tasks.

Suppose continuously differentiable function $h(x)\leq 0$ represents a task, and maximizing it as $h(x)\rightarrow0$ means the completion of the task. Let its zero super-level set be $C\!=\!\{x~|~ h(x)\geq 0\}=\{x~|~ h(x)=0\}$. Then, the control aim is to converge $x$ into $C$.
% Let a continuously differentiable function be $h:\mbr^{n_x}\to\mbr$ and a set $C$ be the zero super-level set of $h$ as $C=\left\{x~\middle|~ h(x)\geq 0\right\}$.
The following definition and theorem ensure forward invariance and asymptotically stability of $C$.
\begin{definition}[Control Barrier Function (CBF) \cite{robot_ecology}]
    Given a system in \eqref{eq:dynamics}, and a function $h(x)$ and its zero super-level set $C$. The function $h(x)$ is a \textit{Control Barrier Function} if there exists a Lipschitz continuous extended class-$\mathcal{K}$ function $\gamma$, such that $\sup\{L_f h(x)+ L_g h(x) u\}\geq -\gamma(h(x))$, where $L_fh(x)$ and $L_gh(x)$ denote Lie derivative. % \cite{sastry_nonlinear}.
\end{definition}
\begin{theorem}[Forward invariance \cite{robot_ecology}]
    If $h(x)$ is a CBF, then any Lipschitz continuous controller $u\in\{u ~|~ L_f h(x)+ L_g h(x) u + \gamma(h(x))\geq0\}$ for \eqref{eq:dynamics} renders the zero super-level set $C$ forward invariance and asymptotically stable in $\mbr^{n_x}$. 
    % Namely, 
    % \begin{align*} 
    %     & x(0)\in C\Rightarrow x(t)\in C\ \forall t\geq 0\\ & x(0)\not\in C\Rightarrow x(t)\rightarrow\in C\ as\ t\rightarrow\infty.
    % \end{align*}
\end{theorem}

% Suppose $h(x)\leq 0$ represents a task, and maximizing it as $h(x)\rightarrow0$ means the completion of the task. Let zero super-level set be $C\!=\!\{x~|~ h(x)\geq 0\}=\{x~|~ h(x)=0\}$. Then, the control aim is to converge $x$ into $C$.
% , where $x(t)$ represents the state $x$ at time $t$.

It is shown in \cite{8814594} that the maximization of $h(x)$ while reducing the energy cost is realized by solving the following optimization problem
\begin{align} 
\mathop{\text{minimize}}\limits_{u,\delta} &~ \varepsilon(u) + \delta^2 \\ 
\text{subject to}&~ L_f h(x) + L_g h(x)u \geq -\gamma (h(x))-\delta. \notag
\end{align}

% \begin{remark}\label{rmk:energy_cost}
%     The existing work reduces energy consumption by minimizing $\|u\|^2$. However, this does not necessarily mean minimizing energy. For instance, an airplane is operated more energy efficiently at high speeds than at low speeds because it can utilize aerodynamic lift. Therefore, in this paper we consider as energy cost function a more general convex function of $u$, denoted by $\varepsilon(u)$.
% \end{remark}

Consider the virtual robot $\mu(i,k)$ now in a multi-task situation. Each task is encoded by a CBF, $h_j, j\in\mathcal{N}_t$. Then, the constrained optimization problems can be combined into a single optimization problem as follows \cite{gennaro2022resilient}:
\begin{align} 
\mathop{\text{minimize}}\limits_{u_{i,k},\delta_{\mu(i,k)}} &~ \varepsilon_{i,k}(u_{i,k}) + \| \delta_{\mu(i,k)} \|^2 \notag\\ 
\text{subject to}&~ L_{f_{i,k}}h_j(x_i) + L_{g_{i,k}}h_j(x_i)u_{i,k} \notag \\
&~\quad\geq -\gamma (h_j(x_i))-\delta_{\mu(i,k)j},~\forall j \in \mathcal{N}_t, \notag
\end{align}
where $\delta_{\mu(i,k)} = [\delta _{{\mu(i,k)}1}\dots\delta _{{\mu(i,k)}n_t}]^\top$ denotes the slack variables for the all tasks.

\subsection{Task Allocation for Multi-mode Robots} 
Now, we are ready to formulate an optimization problem that allocates tasks to multi-mode robots.
Let $\alpha \in \{ 0,1\} ^{n_t\times n_{vr}}$ be an allocation matrix such that
\begin{align}
    \alpha_{j,\mu(i,k)} = \begin{cases}
        1 & \text{if task $j$ is assigned to virtual robot $\mu(i,k)$} \\
        0 & \text{otherwise}
    \end{cases} \notag
\end{align}
We need to ensure no two virtual instances of the same robot---corresponding to its modes---are assigned tasks simultaneously, which means the following condition must be satisfied.
\begin{align}
    \sum_{k=1}^{m_i} \sum_{j=1}^{n_t} \alpha_{j,\mu(i,k)} \leq 1, \quad \forall i\in\mathcal{N}_r \label{eq:summation_alpha}
\end{align}
Considering this condition, the optimization problem for task allocation to multi-mode robots is defined as follows.
\begin{subequations}
\label{eq:task_allocation}
\begin{align}
    %& \text {Task allocation optimization problem (MIQP)} \notag\\ 
    \mathop{\text{minimize}}\limits_{u,\delta,\alpha } &~ \sum _{i = 1}^{n_r}\sum _{k = 1}^{m_i} \left(l_1 \| \Pi _{\mu(i,k)} \alpha_{-,\mu(i,k)}\| ^2 \right.\notag\\
    &\hspace{4em} + \left.\varepsilon_{i,k}(u_{i,k}) + l_2 \| \delta_{\mu(i,k)} \| _{S_{\mu(i,k)}}^2 \right) \label{eq:objective}\\ 
    \text{subject to}&~ L_{f_{i,k}}h_j(x_i) + L_{g_{i,k}}h_j(x_i)u_{i,k} \notag\\
    &~\hspace{5em}\geq -\gamma (h_j(x_i))-\delta_{\mu(i,k)j} \label{eq:task_constraint}\\ 
    &~\Theta_i \delta_{[i]} + \Phi_i \alpha_{[i]} \leq \Psi_i \label{eq:alphas_deltas}\\
    % &~ \sum_{k=1}^{n_m} \sum_{j=1}^{n_t} \alpha_{j,\mu(i,k)} \leq 1 \label{eq:alpha_sum_up}\\
    &~ \mathbbm{1}_{n_tm_i}^\top \alpha_{[i]} \leq 1 \label{eq:alpha_sum_up}\\
    &~ F \alpha _{j,-}^\top\geq T_{j,-}^\top \label{eq:robot_requirement_for_task}\\
    % &~ n_{j,\text{min}} \leq \sum_{i=1}^{n_r} \sum_{k=1}^{n_m} \alpha_{j,\mu(i,k)} \leq n_{j,\text{max}} \label{eq:robot_limits_for_task}\\
    &~ n_{j,\text{min}} \leq \mathbbm{1}_{n_{vr}}^\top \alpha_{j,-}^\top \leq n_{j,\text{max}} \label{eq:robot_limits_for_task}\\
    &~\| \delta_{\mu(i,k)} \|_\infty \leq \delta _\text{max} \label{eq:delta_range} \\
    &~\alpha \in \{ 0,1\} ^{n_t\times n_{vr}} \label{eq:alpha_range} \\
    &~\hspace{4em} \forall k \in \mathcal{N}_{m_i},~\forall j \in \mathcal{N}_t,~\forall i \in \mathcal{N}_r, \notag
\end{align}
\end{subequations}
where $\delta_{[i]}=[\delta_{\mu(i,1)}^\top\dots\delta_{\mu(i,m_i)}^\top]^\top$ and $\alpha_{[i]}=[\alpha_{-,\mu(i,1)}^\top\dots\alpha_{-,\mu(i,m_i)}^\top]^\top$, and the other undefined symbols will be defined hereafter. The program in \eqref{eq:task_allocation} is a mixed-integer quadratic program.

The objective function \eqref{eq:objective}, with weighting parameters $l_1,~l_2\in\mbr_{>0}$, aims to minimize the penalty of task allocation, energy cost, and slack variables. 
The first term prevents tasks from being assigned to infeasible virtual robots using the penalty matrix $\Pi _{\mu(i,k)}$. The second term minimizes the energy costs. The third term minimizes the weighted norm of slack variables with a measure $S_{\mu(i,k)}$. This weighting can relax the constraint on the slack variable of a task with an infeasible virtual robot.
% constraints of task execution 
The constraint \eqref{eq:task_constraint} ensures task execution as introduced in Section \ref{sss:task_as_constraint}.
% constraints of prioritizing
The constraint \eqref{eq:alphas_deltas} is a prioritization between modes and tasks. The matrices $\Theta_i,~\Phi_i\in \mathbb {R}^{\frac{n_t m_i(n_t-1)(m_i-1)}{2}\times n_t m_i}$ and $\Psi_i\in \mathbb {R}^{\frac{n_t m_i(n_t-1)(m_i-1)}{2}}$ are obtained by collecting the inequalities:
\begin{subequations}
\begin{align} 
\delta _{j,\mu(i,k)} \leq \kappa^{-1} \delta _{j,\mu(i,k')}+\delta _\text{max}(1 - \alpha_{j,\mu(i,k)}), \label{eq:mode_priproty}\\
\delta _{j,\mu(i,k)} \leq \kappa^{-1} \delta _{j',\mu(i,k)}+\delta _\text{max}(1 - \alpha_{j,\mu(i,k)}), \label{eq:task_priproty}\\
% \delta _{j,\mu(i,k')} \geq \kappa \left(\delta _{j,\mu(i,k)} - \delta _\text{max}(1 - \alpha_{j,\mu(i,k)}) \right), \label{eq:mode_priproty}\\
% \delta _{j',\mu(i,k)} \geq \kappa \left(\delta _{j,\mu(i,k)} - \delta _\text{max}(1 - \alpha_{j,\mu(i,k)}) \right),\label{eq:task_priproty}\\
\forall k'\in\mathcal{N}_{m_i}\backslash \{k\},~\forall j'\in\mathcal{N}_t\backslash \{j\},\notag 
\end{align}    
\end{subequations}
with $\kappa\gg1$. 
Inequality \eqref{eq:mode_priproty} prioritizes between modes. 
When task $j$ is assigned to robot $i$ in mode $k$, the second term on the right-hand side of \eqref{eq:mode_priproty} becomes zero. This means the slack variable for mode $k$ needs to be smaller than $1/\kappa$ times one for mode $k'$, prioritizing task execution represented in \eqref{eq:task_constraint} of mode $k$ over mode $k'$.
Otherwise, this term becomes $\delta _\text{max}$, which makes the constraint redundant with \eqref{eq:delta_range}.
Analogously, inequality \eqref{eq:task_priproty} prioritizes tasks.

% This is a most weak constraint since $\delta _{j,\mu(i,k)}$ is bounded to $\delta_\text{max}$ by \eqref{eq:delta_range}. 

% When task $j$ is assigned to mode $k$ of robot $i$, \eqref{eq:mode_priproty} becomes 
% \begin{align*}
%     \delta _{j,\mu(i,k)} \leq \frac{1}{\kappa} \delta _{j,\mu(i,k')},~ \forall k'\in\mathcal{N}_{m_i}\backslash \{k\},
% \end{align*}
% which means the slack variable for mode $k$ needs to be smaller than $1/\kappa$ times one for mode $k'$, prioritizing task execution of mode $k$ over task $k'$.
% Otherwise, \eqref{eq:mode_priproty} becomes 
% \begin{align*}
%     \delta _{j,\mu(i,k)} \leq \delta _\text{max} + \frac{1}{\kappa} \delta _{j,\mu(i,k')},~ \forall k'\in\mathcal{N}_{m_i}\backslash \{j\}. \label{eq:delta_notask}
% \end{align*}
% This is a most weak constraint since $\delta _{j,\mu(i,k)}$ is bounded to $\delta_\text{max}$ by \eqref{eq:delta_range}. 
\begin{remark}
    This prioritization aims to switch modes based on energy cost. The modes that can satisfy the task constraint \eqref{eq:task_constraint} with low energy input should be prioritized. If a less efficient mode is prioritized, it will try to satisfy the constraint \eqref{eq:task_constraint} with inefficient input, causing a large energy cost. The solution $\alpha$, which minimizes the total cost, including energy, will be selected (see also discussion in \cite {8795895}).
\end{remark}
% Inequality \eqref{eq:task_priproty} prioritizes tasks and can be interpreted analogously to \eqref{eq:mode_priproty}.

% constraints of the number of robots
Inequalities \eqref{eq:alpha_sum_up}, \eqref{eq:robot_requirement_for_task}, and \eqref{eq:robot_limits_for_task} are constraints for the number of robots.
The constraint \eqref{eq:alpha_sum_up} ensures that no more than one task is assigned to each robot, which is equivalent to \eqref{eq:summation_alpha}.
Inequality \eqref{eq:robot_requirement_for_task} ensures that the required features are assigned to a task.
Finally, via \eqref{eq:robot_limits_for_task} we can set the minimum and maximum number of robots for each task.

% Constraints \eqref{eq:delta_range} and \eqref{eq:alpha_range} constrain $\delta$ to be bounded on $\delta_\text{max}$, and $\alpha$ to be a binary value.

\subsection{Analysis of Convergence}
\label{sss:convergenceanalysis}
This section provides a condition for the optimization problem \eqref{eq:task_allocation} to complete tasks and generate a stable task allocation.

As the task allocation problem for multi-mode robots \eqref{eq:task_allocation} is formulated resembles one proposed in \cite{gennaro2022resilient}, adopting the convergence condition in \cite{gennaro2022resilient} is facilitated. We only need to redefine variables to fit our multi-mode framework as follows.
Let $h(x)=[h_1(x)\dots h_{n_t}(x)]^\top$ with $x=[x_1^\top\dots x_{n_r}^\top]^\top$, $u= [u_{1,1}^\top\dots u_{1,m_i}^\top\dots u_{n_r,1}^\top\dots u_{n_r,m_{n_r}}^\top]^\top$, $\delta = [\delta _{[1]}^\top \dots \delta _{[n_r]}^\top]^\top$, and $\bar{\alpha }= [\alpha _{[1]}^\top \dots \alpha _{[n_r]}^\top]^\top$. 
Then, define $\varphi=[\gamma 
(h(x))^\top~u^\top~\delta^\top~\bar{\alpha}^\top~1]^\top$. 
Additionally, let
$\bar{\Phi } = \mathrm{diag_b}(\Phi_1,\dots,\Phi_{n_r})$, 
$\bar{\Theta }= \mathrm{diag_b}(\Theta_1,\dots,\Theta_{n_r})$, 
and $\bar{\Psi } = \mathrm{diag_b}(\Psi_1,\dots,\Psi_{n_r})$,
where $\mathrm{diag_b}$ returns a block diagonal matrix.
The constraints \eqref{eq:alpha_sum_up}-\eqref{eq:delta_range} can be rewritten as
$\bar{\alpha }^\top A_\alpha ^\top A_\alpha \bar{\alpha } \leq \bar{\alpha }^\top A_\alpha ^\top b_\alpha$ and
$\delta ^\top A_\delta ^\top A_\delta \delta  \leq \delta ^\top A_\delta ^\top b_\delta$.

Then, the following proposition provides a condition to establish the convergence of the task allocation problem, which is represented as a linear matrix inequality (LMI).
\begin{proposition}\label{prp:convergence}
    If there exist positive scalars $\tau_1$, $\tau_2$, and $\tau_3$ for all time $t\geq0$ that satisfy
    \begin{align}
        B_0 \leq \tau _1 B_1 + \tau _2 B_2 + \tau _3 B_3,
    \end{align}
    where $B_0$, $B_1$, $B_2$, and $B_3$ with $c\in\mbr_{>0}$ are given in Appendix, then the sequences $u(t),~\delta(t),~\alpha(t)$, and solutions of \eqref{eq:task_allocation} converge as $t\to\infty$.
\end{proposition}
\begin{proof}
    Similar to \cite{gennaro2022resilient}.
\end{proof}

% \begin{remark}
%     \red{This proposition is useful for situations where all robots can execute all tasks.
%     Otherwise, we can add a weight matrix to a Lyapunov function based on the priority.}
% \end{remark}

% \begin{example}[Example for 1D case]
% Here, I consider a robot with a single input and single state system.
% There are two dynamics as
% \begin{align}
%     &\dot{x} = u, &f_1(x) = 0,~g_1(x) = 1\\
%     &\dot{x} = -kx +u, &f_2(x) = -kx,~g_2(x) = 1
% \end{align}
% I set the start position at $x=-4$, and the task is minimizing a distance to a goal point $x=2$ as
% \begin{align}
%     h_1 = h_2 = -(x-2)^2
% \end{align}
% Then, the results are shown in Fig. \ref{fig:1d_switch} and \ref{fig:1d_switch_l2}.
% \end{example}

% \begin{figure}[t]
%     \centering
%     \includegraphics[width=\linewidth]{figure/multimode/1d_mode_switch_sim.pdf}
%     \caption{$l=10^{-2}$}
%     \label{fig:1d_switch}
% \end{figure}

% \begin{figure}[t]
%     \centering
%     \includegraphics[width=\linewidth]{figure/multimode/1d_mode_switch_sim_l2.pdf}
%     \caption{$l=10^2$}
%     \label{fig:1d_switch_l2}
% \end{figure}

\section{HIGH RELATIVE DEGREE TASK EXECUTION AND ALLOCATION}\label{sec:highorder}
This section extends the proposed framework to a scenario with task execution constraints that lead to high relative degree. As a motivating example, we utilize a convertible UAV introduced in Example~\ref{ex:convertible_UAV} with tasks represented by its position. In such scenarios, $L_{g_{i,k}}h_j(x_i)$ in \eqref{eq:task_constraint} becomes zero; hence, the constraint \eqref{eq:task_constraint} cannot be employed as a constraint for the input. 
%This section extends the proposed framework to robots with a high relative degree model, using the example of position-related tasks executed by convertible UAVs whose velocity has dynamics. 
%In such a case, using the framework of Section \ref{sec:task_allocation} is not straight because $L_{g_{i,k}}h_j(x_i)$ in \eqref{eq:task_constraint} becomes zero and does not constrain input to execute tasks.
To address this issue, Section \ref{sss:highorder_execution} and Section \ref{sss:highorder_condition} extend the constraint \eqref{eq:task_constraint} and Proposition \ref{prp:convergence}, respectively.

%% ===========================================
\subsection{Modeling a Convertible UAV}
%% ===========================================
\subsubsection{Dynamics}
The convertible UAVs described in Example~\ref{ex:convertible_UAV} have two modes, 1: cruise and 2: hovering. Each of these two modes has different control inputs that stem from their features. Specifically, the cruise mode is driven by a forward velocity and a yaw rate input, while the hovering mode is governed by horizontal velocity input. 
%Consider convertible UAVs with two modes, 1: cruise and 2: hovering. A difference in the control input distinguishes these two modes. Suppose we can input a forward velocity and a yaw rate in the cruise mode while we can input 2-D velocities in the hovering mode.

Let us first derive a generalized equation of motion shared between both modes. Let $ x_i\in\mbr^2$ and $ \eta_i=[v_{i,x}~v_{i,y}~\theta_i]^\top$ be the position and the input consisting of horizontal velocity and yaw orientation of the UAV $i$, respectively.
Then, the state derivative can be written as
\begin{align}
    \dot{ x_i}
    =\begin{bmatrix}
        \cos(\theta_i) & -\sin(\theta_i)\\
        \sin(\theta_i) & \cos(\theta_i)
    \end{bmatrix}
    \begin{bmatrix}
        v_{i,x} \\ v_{i,y}
    \end{bmatrix} =  f_i( \eta_i) \label{eq:general_ss}
\end{align}
Additionally, to prevent a discontinuous change in the linear velocity and to model the transition phase between the two modes, we introduce the input dynamics as
\begin{align}
    \dot{v}_{i,x} \!=\! k_v(v_{i,x}^\text{ref}\!-\!v_{i,x}),~
    \dot{v}_{i,y} \!=\! k_v(v_{i,y}^\text{ref}\!-\!v_{i,y}),~
    \dot{\theta_i} \!=\! \omega_i,\label{eq:input_dynamics}
\end{align}
% \begin{align}
%     \dot{ \eta}_i = \begin{bmatrix}
%         k_v(v_{i,x}^\text{ref}-v_{i,x}) \\
%         k_v(v_{i,y}^\text{ref}-v_{i,y}) \\
%         \omega_i
%     \end{bmatrix},\label{eq:input_dynamics}
% \end{align}
with positive gain $k_v\in\mbr_{>0}$.
Then, \eqref{eq:input_dynamics} can be rewritten as
\begin{align}
    \dot{ \eta}_i = \begin{bmatrix}
        -k_v v_{i,x} \\
        -k_v v_{i,y} \\
        0
    \end{bmatrix}
    +\begin{bmatrix}
        k_v v_{i,x}^\text{ref} \\
        k_v v_{i,y}^\text{ref} \\
        \omega_i
    \end{bmatrix} = \psi(\eta_i) + g_i\begin{bmatrix}
        v_{i,x}^\text{ref} \\ v_{i,y}^\text{ref} \\ \omega_i
    \end{bmatrix}.\label{eq:general_dynamics}
\end{align} 
Because $\psi(\eta_i)$ can be regarded as a viscous friction, this formulation mitigates a discontinuity in vehicle velocity.

Next, we derive dynamics specific to each mode based on \eqref{eq:general_ss} and \eqref{eq:general_dynamics}. For the cruise mode, its input is composed of a forward velocity and yaw rate as $u_{i,1} = [v_{i,x}^\text{ref}~\omega_i]^\top$. To incorporate this new input into \eqref{eq:general_dynamics}, we substitute $v_{i,y}^\text{ref}=0$ and replace $g_i$ with $g_{i,1}$ which is obtained by eliminating the second column of $g_i$.
%Next, we specialize \eqref{eq:general_ss} and \eqref{eq:general_dynamics} to each mode. 
%For the cruise mode, let new input be $u_{i,1} = [v_{i,x}^\text{ref}~\omega_i]^\top$, substitute $v_{i,y}^\text{ref}=0$, and obtain new matrix $g_{i,1}$ by removing the second column from $g_i$. 
Then, we get the state space equation of the cruise mode as 
\begin{subequations}\label{eq:hybrid_dynamics}
\begin{align}
    &\frac{d}{dt}\begin{bmatrix}
         x_i \\  \eta_i
    \end{bmatrix}
    =\begin{bmatrix}
         f_i( \eta_i) \\ 
         \psi(\eta_i) + g_{i,1}u_{i,1}
    \end{bmatrix}.
\end{align}
By following similar procedures, the state space equation of hovering mode is derived as 
%Similarly, analogous procedures with $u_{i,2} = [v_{i,x}^\text{ref}~v_{i,y}^\text{ref}]^\top$, $\omega_i=0$, and $g_{i,2}$, which obtained by removing the third column from $g_i$, give the state space equation of hovering mode as
\begin{align}
    &\frac{d}{dt}\begin{bmatrix}
         x_i \\  \eta_i
    \end{bmatrix}
    =\begin{bmatrix}
         f_i( \eta_i) \\ 
         \psi(\eta_i) + g_{i,2}u_{i,2}
    \end{bmatrix}.
\end{align}
\end{subequations}
where $u_{i,2} = [v_{i,x}^\text{ref}~v_{i,y}^\text{ref}]^\top$ and $\omega_i=0$. Note that $g_{i,2}$ is obtained by removing the third column from $g_i$. 

%% ===========================================
\subsubsection{Energy model}
%% ===========================================
% As mentioned in Remark \ref{rmk:energy_cost}, minimizing input does not always mean minimizing energy consumption. Therefore, we define the shifted quadratic energy cost functions for cruise and hovering modes as 
We define energy cost functions as
\begin{subequations}\label{eq:energy_cost}
\begin{align}
    &\varepsilon_{i,1}(u_{i,1})= (v_{i,x}^\text{ref} - v_x^\text{eff})^2 + \omega_i^2, \\
    &\varepsilon_{i,2}(u_{i,2})= v_{i,x}^{\text{ref}2} + v_{i,y}^{\text{ref}2},
\end{align}    
\end{subequations}
where $v_x^\text{eff}\in\mbr_{>0}$ is the forward velocity of the cruise mode that minimizes energy cost. 
% If one ignores $v_{i,y}^\text{ref}$ and $\omega_i$, 
If we consider only $v_{i,x}^\text{ref}$, under high forward velocity reference, cruise mode is regarded as more energy-efficient than hovering mode. That means cruise mode has an advantage in executing a task that requires high-velocity forward movement.

% The above quadratic energy cost does not necessarily represent exact energy consumption because it is difficult to model under uncertain aerodynamics. However, the quadratic formulation can approximate it and has advantages in handling and implementation. 

%% ===========================================
\subsection{High Relative Degree Task Execution} \label{sss:highorder_execution}
%% ===========================================
Here, we extend the condition for task execution \eqref{eq:task_constraint} to a high relative degree case.
The task used in the simulations is to minimize the distance between a robot and a target point. Hence, we define a CBF as 
\begin{align}
    h_{j}(x_i) = -\| x_i -  p_j\|^2, \label{eq:h1}
\end{align}
where $p_j$ is the task $j$'s target position. The convergence to its zero super-level set $C_1=\{x_i\mid h_j(x_i)\geq 0\}=\{p_j\}$ means the accomplishment of the task $j$. 

Because the UAV model has input dynamics, we need to employ integral CBFs \cite{9131819}. 
According to \cite{9131819}, $C_1$ is rendered to be forward invariant when the input satisfies 
\begin{align}
    &\dot{h}_j + \gamma_1(h_j) \notag\\
    &\!=\!\frac{\partial h_j}{\partial x_i}f_i(\eta_i) \!+\! \underbrace{\frac{\partial h_j}{\partial \eta_i}}_{=0}\left(\phi_i(\eta_i)\!+\!g_{i,k}u_{i,k}\right)\!+\!\gamma_1(h_j)\geq0, \label{eq:first_CBF}
\end{align} 
where $\gamma_1$ is an extended class-$\mathcal{K}$ function.
However, as $\partial h_j/\partial \eta_i$ is zero, \eqref{eq:first_CBF} does not constrain the actual input $u_{i,k}$. 
Then, we define a second CBF candidate as 
\begin{align}
    h'_j(x_i,\eta_i) = \frac{\partial h_j}{\partial x_i}f_i(\eta_i) +\gamma_1(h_j). \label{eq:h2}
\end{align} 
The input $u_{i,k}$ which satisfies 
\begin{align}
    \dot{h}'_j + \gamma_2(h'_j) =& \frac{\partial}{\partial x_i}\left(\frac{\partial h_j}{\partial x_i}f_i(\eta_i)\right)f_i(\eta_i) \notag\\
    & +\frac{\partial}{\partial \eta_i}\left(\frac{\partial h_j}{\partial x_i}f_i(\eta_i)\right)\left(\phi_i(\eta_i)+g_{i,k}u_{i,k}\right) \notag\\
    & +\frac{\partial \gamma_1}{\partial h_j}\dot{h}_j + \gamma_2(h'_j(x_i)) \geq 0\label{eq:second_CBF}
\end{align}
renders $C_2=\{(x_i,\eta_i)\mid h'_j(x_i,\eta_i)\geq 0\}$ forward invariant; simultaneously, this input also renders $C_1$ forward invariant (see discussion in \cite{robot_ecology}).
Therefore, we can utilize \eqref{eq:second_CBF} in \eqref{eq:task_allocation}, instead of the constraint \eqref{eq:task_constraint}.

\subsection{Convergence of High Relative Degree Task Allocation} \label{sss:highorder_condition}

In this section we give sufficient conditions for the convergence of the task allocation algorithm \eqref{eq:task_allocation} for the case of high relative degree tasks.
\begin{proposition}\label{prp:convergence_2}
    If there exist positive scalars $\tau_1$, $\tau_2$, and $\tau_3$ for all time $t\geq0$ that satisfy
    \begin{align}
        B'_0 \leq \tau _1 B_1 + \tau _2 B_2 + \tau _3 B_3,
    \end{align}
    where $B'_0$, $B_1$, $B_2$, and $B_3$ are reported in the Appendix, then the sequences $u(t)$, $\delta(t)$, $\alpha(t)$, and solutions of \eqref{eq:task_allocation} converge as $t\to\infty$.
\end{proposition}
\begin{proof}
    Let the candidate Lyapunov function be 
    \begin{align}
        V(x) = h(x)^\top h(x). \label{eq:lyapunov}
    \end{align}
    Then, we want the following condition on its time derivative to hold: 
    \begin{equation}
        \dot{V}(x, \eta) = 2h(x)^\top \frac{\partial h}{\partial x}f(\eta) \leq -c_1 V(x).\label{eq:lyapunov_2}
    \end{equation}
    with $c_1\in\mbr_{>0}$, where $f(\eta)=[f_1(\eta)^\top\dots f_{n_r}(\eta)^\top]^\top$ with $\eta=[\eta_1^\top\dots \eta_{n_r}^\top]^\top$. We define the auxiliary Lyapunov function $V'(x, \eta) = -\dot{V}(x, \eta) -c_1 V(x) \geq 0$, and enforce the stability condition $\dot{V}'(x, \eta) \leq -c_2 V'(x, \eta)$, for $c_2\in\mbr_{>0}$, which can be written as
    \begin{equation}
        -\ddot{V}(x, \eta)-(c_1+c_2)\dot{V}(x, \eta)-c_1c_2V(x)\leq 0.\label{eq:ddotV}
    \end{equation}
    The condition \eqref{eq:ddotV} can be further simplified to
    \begin{align}
        \varphi^\top  B'_0 \varphi \leq 0,
    \end{align}
    where $B'_0$ is given in the Appendix and $\varphi$ is given in Section~\ref{sss:convergenceanalysis}.
    The same steps used in proposition \ref{prp:convergence} complete the proof.
\end{proof}

\section{SIMULATIONS}

\begin{figure}
    \centering
    \subfloat[]{\includegraphics[width=0.35\linewidth]{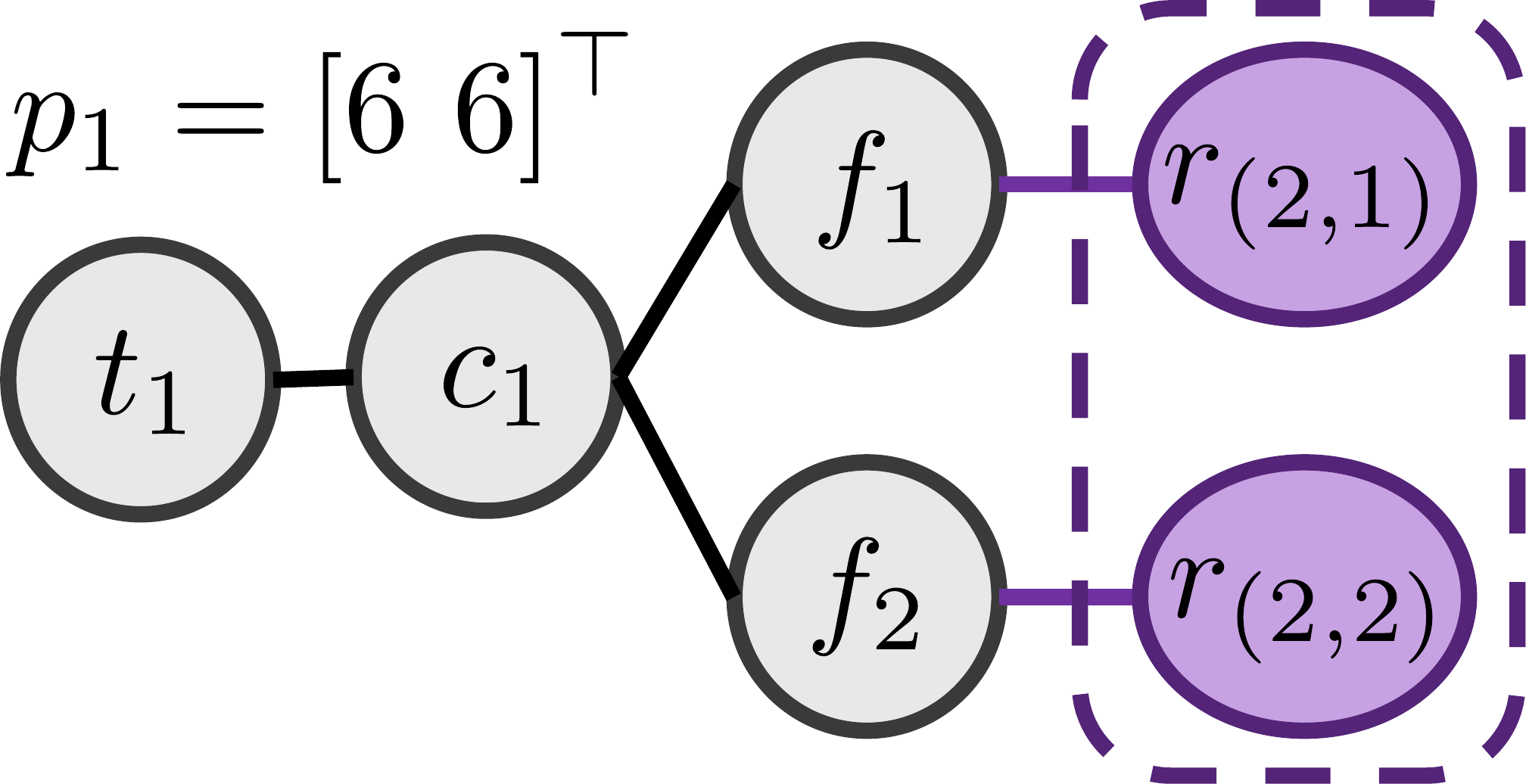}%
    \label{fig:single_encode}}
    \hfil
    \subfloat[]{\includegraphics[width=0.30\linewidth]{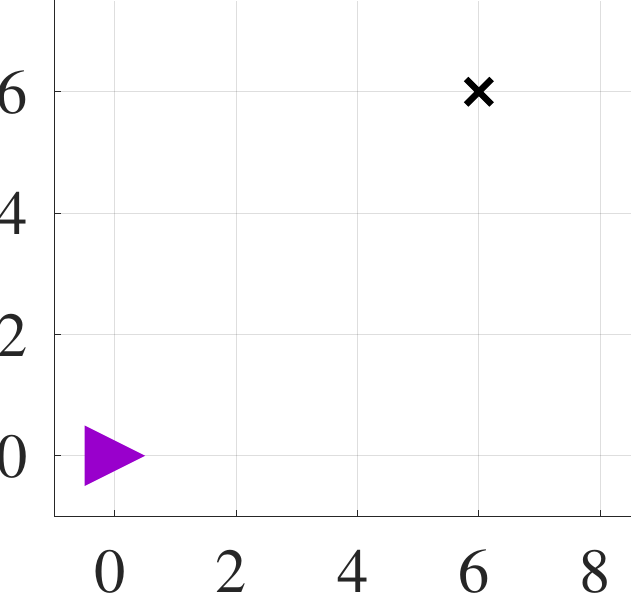}%
    \label{fig:single_init}}
    \hfil
    \subfloat[]{\includegraphics[width=0.30\linewidth]{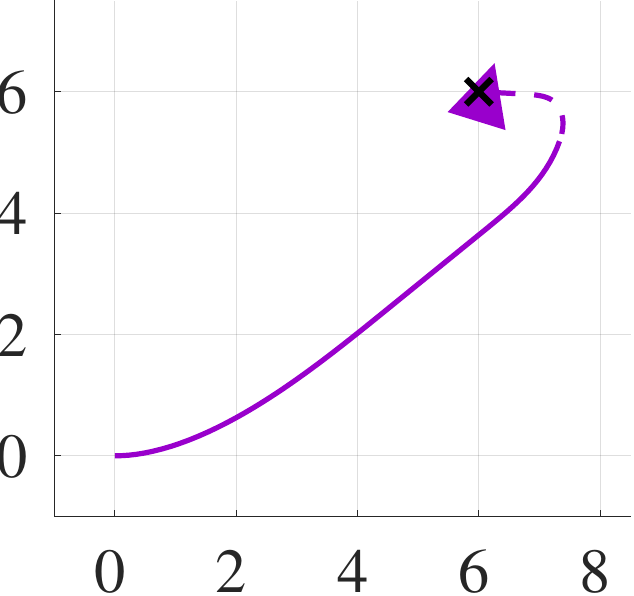}%
    \label{fig:single_result}}
    \caption{Single UAV simulation. (a) Encoding: Both modes can execute the task. (b) Initial state. (c) Result trajectory: The trajectories of the cruise and hovering mode are shown as solid and dashed lines, respectively.}
    \label{fig:single_sim}
\end{figure}

% \begin{figure}
%     \centering
%     \subfloat[]{\includegraphics[width=0.35\linewidth]{figure/simulation/multisim_encode.pdf}%
%     \label{fig:multi_encode}}
%     \hfil
%     \subfloat[]{\includegraphics[width=0.32\linewidth]{"figure/simulation/bandmud_0.pdf"}%
%     \label{fig:band_mud0}}
%     \hfil
%     \subfloat[]{\includegraphics[width=0.32\linewidth]{"figure/simulation/bandmud_1.pdf"}%
%     \label{fig:band_mud1}}
%     \hfil
%     \subfloat[]{\includegraphics[width=0.32\linewidth]{"figure/simulation/bandmud_2.pdf"}%
%     \label{fig:band_mud2}}
%     \hfil
%     \subfloat[]{\includegraphics[width=0.32\linewidth]{"figure/simulation/bandmud_3.pdf"}%
%     \label{fig:band_mud3}}
%     \hfil
%     \subfloat[]{\includegraphics[width=0.32\linewidth]{"figure/simulation/bandmud_10.pdf"}%
%     \label{fig:band_mud10}}
%     \caption{Multiple UAV simulation with a regulation restriction. Each robot is drawn in the same color in all figures. (a) Encoding: Robot 1 has only cruise mode, while the others have both modes. (b) Snapshot of the result at the time $t=0$. The result trajectories of the cruise and hovering modes are drawn in solid and dashed lines in the same color of its robot. The cruise mode is prohibited in the brown area. (c) $t=1$. (d) $t=2$. (e) $t=3$. (f) $t=10$.}
%     \label{fig:band_mud}
% \end{figure}

\begin{figure*}
    \centering
    \subfloat[]{\includegraphics[width=0.18\linewidth]{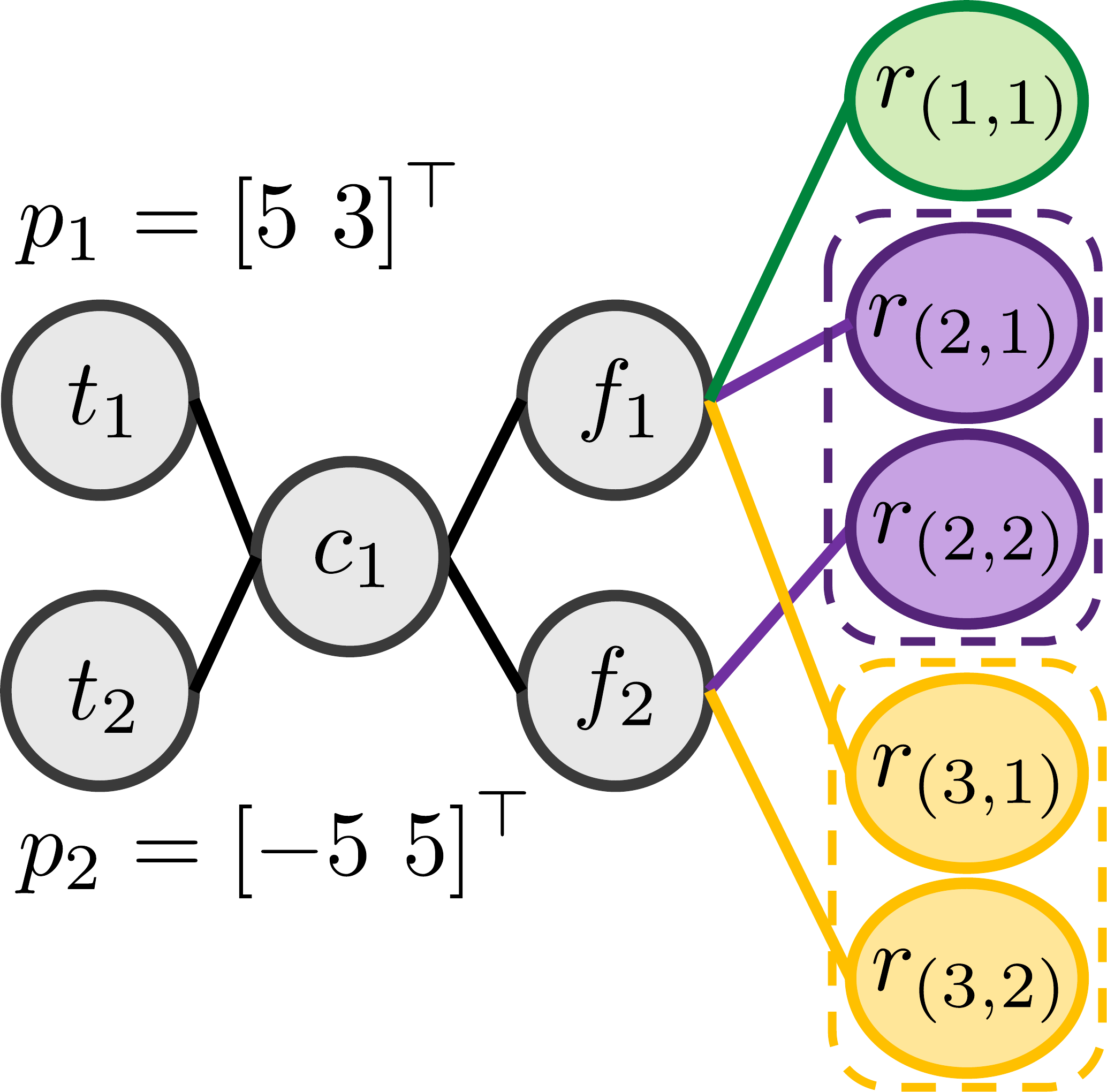}%
    \label{fig:multi_encode}}
    \hfil
    \subfloat[]{\includegraphics[width=0.16\linewidth]{"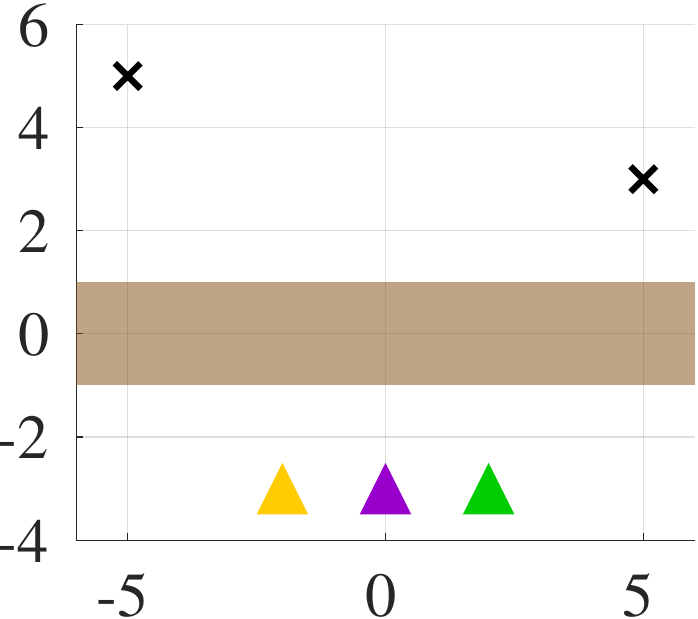"}%
    \label{fig:band_mud0}}
    \hfil
    \subfloat[]{\includegraphics[width=0.16\linewidth]{"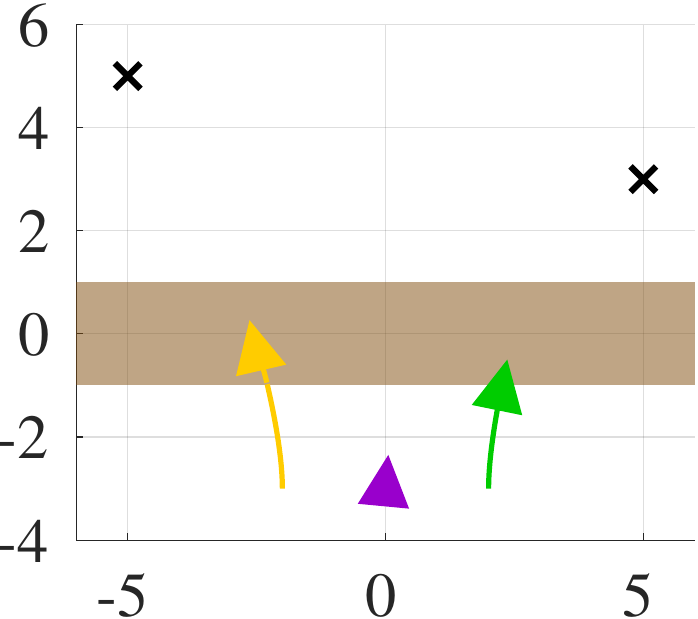"}%
    \label{fig:band_mud1}}
    \hfil
    \subfloat[]{\includegraphics[width=0.16\linewidth]{"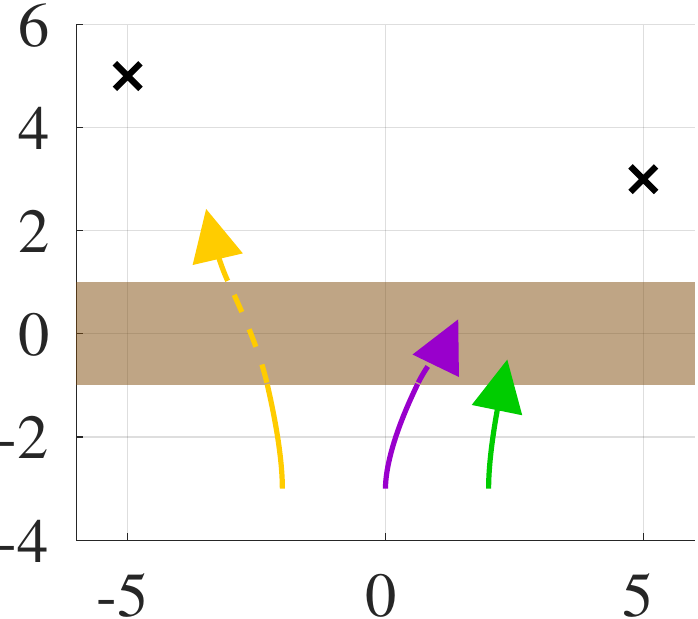"}%
    \label{fig:band_mud2}}
    \hfil
    \subfloat[]{\includegraphics[width=0.16\linewidth]{"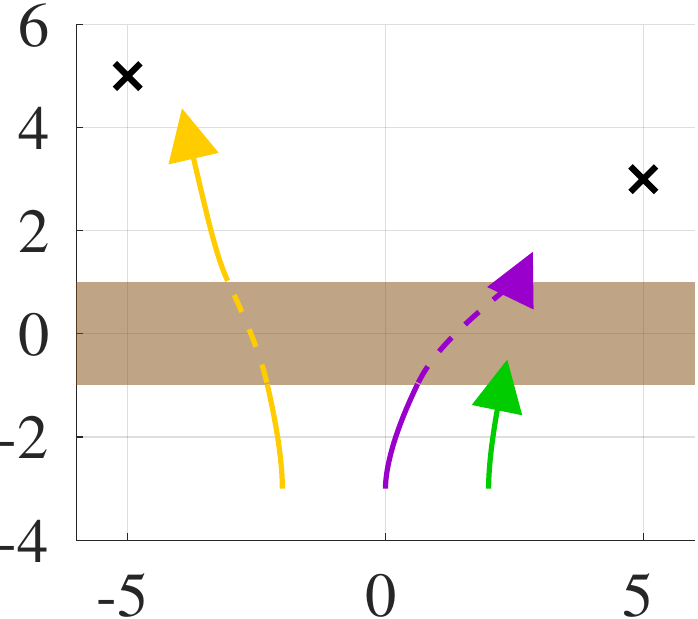"}%
    \label{fig:band_mud3}}
    \hfil
    \subfloat[]{\includegraphics[width=0.16\linewidth]{"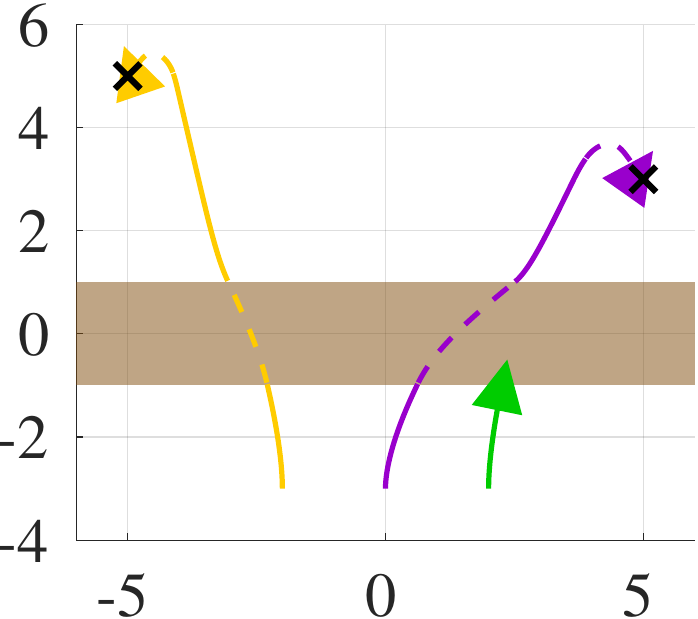"}%
    \label{fig:band_mud10}}
    \caption{Multiple UAV simulation with a regulation restriction. Each robot is drawn in the same color in all figures. (a) Encoding: Robot 1 has only cruise mode, while the others have both modes. (b) Snapshot of the result at the time $t=0$ s. The result trajectories of the cruise and hovering modes are drawn in solid and dashed lines in the same color of its robot. The cruise mode is prohibited in the brown area. (c) $t=1$ s. (d) $t=2$ s. (e) $t=3$ s. (f) $t=10$ s.}
    \label{fig:band_mud}
\end{figure*}

This section provides simulations to demonstrate the proposed method with the settings in Section \ref{sec:highorder}. The parameters are set as $k_v=4$, $l_1=10^6$, $l_2=10^{-4}$, $\kappa=10^4$, $\delta_\text{max}=10^4$, $v_x^\text{eff}=2$ m/s, $\gamma_1(h)=5h$, and $\gamma_2(h)=h$.
We conduct two simulations: one is a single task with a single UAV, and the other is multiple tasks with multiple UAVs under the constraint that cruise mode is prohibited in a given area (modeling, e.g., a regulatory restriction).

The graph encoding the first simulation is shown in Fig.~\ref{fig:single_sim}~\subref{fig:single_encode}. The initial state and the target position of the task are shown in Fig.~\ref{fig:single_sim}~\subref{fig:single_init}. 
The result trajectory is shown in Fig.~\ref{fig:single_sim}~\subref{fig:single_result}. Initially, when the robot flies forward, the cruise mode is assigned to save energy. Then, when the robot gets closer to the target point, the hovering mode is assigned to accomplish the task. This result shows the proposed allocation method can switch robot modes according to energy consumption and task execution ability.

The second simulation features multiple UAVs and tasks. The initial state and the target points of the tasks are set as shown in Fig.~\ref{fig:band_mud} \subref{fig:band_mud0}, where the crosses represent the task target points. Moreover, we enforce a no-cruise area to model a regulatory restriction shown as the brown area in Fig.~\ref{fig:band_mud}. If a UAV enters this area, the specialization matrix of the cruise mode $S_{\mu(i,1)}$ becomes a zero matrix, preventing the tasks from being assigned to cruise mode. There are three UAVs, and the rightmost UAV can only cruise; therefore, it would immediately stop as it enters the restricted area. The remaining UAVs are able to both cruise and hover.
Fig.~\ref{fig:band_mud} shows the snapshots of the second simulation result. At first, the nearest UAVs are assigned to each task---an allocation that optimizes energy consumption (Fig.~\ref{fig:band_mud} \subref{fig:band_mud0}-\subref{fig:band_mud1}). However, at time $t=1$ s (Fig.~\ref{fig:band_mud} \subref{fig:band_mud1}), the right UAV stops because of the regulation; then, the middle UAV, which has not been assigned to any task, is recruited to go to the rightmost cross. Finally, the two UAVs complete each task by switching their modes according to the regulation and energy consumption (Fig.~\ref{fig:band_mud} \subref{fig:band_mud2}-\subref{fig:band_mud10}).
This result demonstrates that the proposed method provides resilience under environmental restrictions.

\section{CONCLUSION}
This work proposed a novel multi-robot task allocation framework designed for multi-mode robots. First, the multimodality was encoded as a linear mapping between tasks and modes, where robot modes are represented by virtual robot nodes. Then, a constrained optimization problem was formulated to allocate tasks considering robot capabilities, energy consumption, and task execution. The proposed framework is able to encompass high relative degree task models, and guarantees of task allocation convergence are provided. 
Simulations demonstrated how the proposed framework allows robots to execute tasks by switching their mode according to energy consumption. 
%Additionally, it was verified that the proposed method can reallocate tasks even when some robots stop under certain environmental restrictions.

\section*{APPENDIX}\label{apdx:matrices}

The matrices used in Proposition \ref{prp:convergence} and \ref{prp:convergence_2} are defined as symmetric sparse matrices whose upper triangular parts are defined as follows, where $B_*(i,j)$ indicates $(i,j)$ block of the matrix.
$B_0(1,1)=cI$, 
$B_0(1,3)=\frac{{d}\gamma }{{d}h}\frac{{d}h}{{d}x} g(x)$, 
$B_0(1,5)=\frac{{d}\gamma }{{d}h}\frac{{d}h}{{d}x} f(x)$. $c\in\mbr_{>0}$.
$B_1(1,3)=-\frac{1}{2}I$, 
$B_1(2,3)=-\frac{1}{2}L_gh(x)^\top$, 
$B_1(3,3)=-I$, 
$B_1(3,5)=-\frac{1}{2}L_fh(x)$.
$B_2(3,4)=\frac{1}{2}\bar{\Theta }^\top \bar{\Phi }$, 
$B_2(4,4)=\bar{\Phi }^\top \bar{\Phi }$, 
$B_2(4,5)=-\frac{1}{2}\bar{\Phi }^\top \bar{\Psi }$.
$B_3(3,3)=A_\delta ^\top A_\delta$, 
$B_3(3,5)=-\frac{1}{2}A_\delta ^\top b_\delta$, 
$B_3(4,4)=A_\alpha ^\top A_\alpha$, 
$B_3(4,5)=-\frac{1}{2}A_\alpha ^\top b_\alpha$.
$B'_0(1,1)=c_1c_2 I$, 
$B'_0(1,2)=\frac{\partial}{\partial \eta}\!\left(\frac{\partial h}{\partial x}f(\eta)\right)g$, 
$B'_0(1,5)=\frac{\partial}{\partial x}\!\left(\frac{\partial h}{\partial x}f(\eta)\right)\!f(\eta)\!+\!\frac{\partial}{\partial \eta}\!\left(\frac{\partial h}{\partial x}f(\eta)\right)\!\phi(\eta)\!+\!(c_1\!+\!c_2)\frac{\partial h}{\partial x}f(\eta)$, 
$B'_0(5,5)\!=\!-2 \left\| \frac{\partial h}{\partial x}f(\eta)\right\|^2$. $c_1,c_2\in\mbr_{>0}$.

% \section*{ACKNOWLEDGMENT}

\bibliographystyle{IEEEtran}
\bibliography{biblio}

\end{document}